\documentclass[11pt,a4paper]{article}
\usepackage[hyperref]{AutoLaw}
\usepackage{times}
\usepackage{tabularx}
\usepackage{graphicx}
\usepackage{multirow}
\usepackage{latexsym}

\usepackage{microtype}

\aclfinalcopy 

\title{AutoLAW: Augmented Legal Reasoning through Legal Precedent Prediction}

\author{Robert Zev Mahari \\
  MIT Media Lab \\
  Harvard Law School \\
  }

\date{}

\begin{document}
\maketitle
\begin{abstract}

This paper demonstrate how NLP can be used to address an unmet need of the legal community and increase access to justice.
The paper introduces Legal Precedent Prediction (LPP), the task of predicting relevant passages from precedential court decisions given the context of a legal argument.
To this end, the paper showcases a BERT model, trained on 530,000 examples of legal arguments made by U.S. federal judges, to predict relevant passages from precedential court decisions given the context of a legal argument.
In 96\% of unseen test examples the correct target passage is among the top-10 predicted passages.
The same model is able to predict relevant precedent given a short summary of a complex and unseen legal brief, predicting the precedent that was actually cited by the brief's co-author, former U.S. Solicitor General and current U.S. Supreme Court Justice Elena Kagan.

\end{abstract}

\section{Introduction}

\begin{table*}[ht]
\centering
\begin{tabular}{|p{0.3\linewidth} | p{0.65\linewidth}|}
\hline
\textbf{Case Background} & [Is] a live, human-made micro-organism patentable subject matter[?]  \\
\hline
\textbf{Legal Question} & We must determine whether respondent’s microorganism constitutes a “manufacture” or “composition of matter” within the meaning of the statute.\\
\hline
\textbf{Legal Standard} & \textbf{“unless otherwise defined, words will be interpreted as taking their ordinary, contemporary, common meaning.”} \textit{Perrin v. United States}, 444 U. S. 37, 42 (1979). \\
\hline
\textbf{Standard Application} & This Court [defines] \textbf{``manufacture” as ``the production of articles for use from raw or prepared materials by giving to these materials new forms, qualities, properties, or combinations, whether by hand-labor or by machinery."} \textit{American Fruit Growers, Inc. v. Brogdex Co.}, 283 U. S. 1, 11 (1931)... The patentee has produced a new bacterium with markedly different characteristics from any found in nature... \\
\hline
\textbf{Conclusion} & [The patentee's] discovery is not nature’s handiwork, but his own; accordingly it is patentable subject matter \\
\hline
\end{tabular}
\caption{Anatomy of a typical legal argument: In \textit{Diamond v. Chakrabarty}, 447 U.S. 303 (1980) the U.S. Supreme Court cites two passages of precedent (bold). The first expresses a legal standard concerning statutory interpretation, the second provides a definition of ``manufacture". The Court applies the standard to conclude that the creation of new micro-organisms constitutes ``manufacture" and is therefore patentable. In this instance, an LPP model should predict the passage from \textit{Perrin} cited in the Legal Standard,  given the Case Background, Legal Question, Standard Application, and Conclusion above. }
\label{tab:common_law}
\end{table*}

In recent years, the law has increasingly attracted attention from the NLP community. \citep{zhong-etal-2020-nlp}.
Several of the recently proposed legal NLP methods solve  technical NLP challenges unique to the law but fail to address the needs of the legal community.
For example, Legal Judgement Prediction (LJP) seeks to predict a legal judgement (ordinarily made by judges) on the basis of the relevant facts and laws \citep{,aletras2016predicting,luo2017learning,zhong2018legal, chen2019charge,chalkidis-etal-2019-neural}.
In practice however, judges are unlikely to defer to artificial intelligence to decide the fate of a case.
Meanwhile, members of the legal community have grown skeptical about NLP, suggesting that NLP ``has introduced uncertainty to the law" \citep{callister2020law}.

The intention of this paper is to introduce Legal Precedent Prediction (LPP) as a new NLP task which addresses an unmet need of the legal community.
LPP will be defined as the task of predicting passages of judicial precedent that are relevant to a given legal argument made in the context of a judicial opinion or a legal brief, and might provide an appropriate citation for a given proposition in the opinion or brief.
LPP has the potential to promote access to justice by augmenting attorneys' ability to identify precedent in a cost- and time-efficient manner and thereby reduce the cost of litigation.

In common law jurisdictions, like the United States, judges and lawyers construct their legal arguments by drawing on judicial precedent from prior opinions (Table \ref{tab:common_law}).
Judges cite precedent in their opinions and apply it to the facts of a case to build incrementally towards a final judgement.
Lawyers use precedent in the legal briefs they present to courts to argue why one party to the case should prevail. 
Legal briefs are structured similarly to judicial opinions but advocate for a certain legal conclusion (e.g., the defendant's actions cannot be considered a ``crime" under current law).
Both judicial opinions and legal briefs usually contain a number of independent legal arguments - each citing its own set of precedent.
The precedent contained in these arguments depends on the context of the entire case as well as on the specific legal argument being made.

U.S. case law currently consists of around 6.7 million published judicial opinions, written over 350 years \footnote{Based on the Case Law Access Project (CAP) corpus.}.
The process of extracting the correct precedent from this daunting corpus is a fundamental part of legal practice.
It is estimated that law firm associates spend one-third of their working hours conducting legal research \citep{lastres2015rebooting}.
Lawyers rely on legal research platforms to access and search legal precedent, which charge \$60 to \$99 per search, a cost that is ordinarily passed on to clients \citep{LegalResearchPricing}.

Access to justice continues to be a serious problem in the United States, and ``86\% of the civil legal problems reported by low-income Americans received inadequate or no legal help" \citep{legal2017justice}.
Similarly, in criminal cases, U.S. public defenders, who provide legal counsel to individuals who have been charged with a crime and face imprisonment but cannot afford a lawyer, frequently handle hundreds of cases simultaneously and are thus unable to devote the necessary time to each individual's case \citep{oppel2019one}.
As attorney fees continue to rise and approach \$300 per hour on national average \citep{ClioLegalTrends} the price of legal advice is becoming increasingly unaffordable and access to justice is diminishing accordingly.

Identifying precedent is a task fundamental to the practice of law.
Given the time, expertise, and costs associated with identifying relevant precedent, this task represents a major barrier for widespread access to justice.
This paper presents an NLP approach to predicting judicial precedent relevant to a given legal argument by training a model on legal arguments made by U.S. federal judges in judicial opinions.
The goal is for such a model to aid attorneys in drafting legal briefs, reducing time and money spent on legal research and ultimately increasing access to justice.

\section{Related Work}
To the best of my knowledge, \citet{dadgostari2020modeling} present the only published approach to predicting legal citations.
The authors formulate this task as an information retrieval problem; they use a bag-of-words representation of a given judicial opinion to model the opinion's topic and search the ``topic space" for related opinions.
\citet{dadgostari2020modeling} train their models on U.S. Supreme Court opinions.
The authors evaluate their model by predicting numerous citations relevant to a given opinion and checking how many of these are actually contained in the opinion, achieving a Precision@10 of 19\%.

The LPP model presented here seeks to identify a specific passage from prior opinions relevant to a specific legal argument and achieves a precision of 72\%.
The LPP model is trained on judicial arguments in opinions from all 108 courts in the U.S. federal court system and is  tested on judicial opinions and legal briefs.
By training the LPP model on individual judicial arguments, the model can learn domain specific connections that a topic model alone might miss (e.g. the foundational U.S. property law case \textit{Pierson v. Post}, 2 Am. Dec. 264 (1805) revolves around hunting a wild fox). 

\section{Methodology}

\subsection{Data}

The Case Law Access Project (CAP) provides researchers with access to 6.7 million published judicial opinions.

In these opinions, judges cite precedent by quoting directly, summarizing, or simply referencing precedent in support of an argument.
A single opinion is likely to contain many arguments, each resting on a different set of citations.

This paper focuses on the 1.7 million federal judicial opinions contained in CAP which include all opinions handed down from the U.S. Supreme Court, 13 federal appellate courts and 94 federal district courts.
These 1.7 million opinions contain 13.8 million citations of precedent, 7.4 million of which are accompanied by a quoted passage from a prior case.
Regular expressions were used to extract these citation-passage pairs and match them to the cited opinion text (to exclude any inaccurate citations).
Ultimately, 1.5 million unique cited passages were identified.

Judicial citations obey a power law distribution (Figure \ref{fig:ccdf}) .
The 5,000 most frequently cited passages were selected to train the LPP model.
Although these passages represent less than 0.5\% of all cited passages, they account for 19\% of all passage citations and, as will be shown, appear frequently in legal briefs.
These frequently cited passages appear a total of 560,000 times.

\subsection{Problem Formulation and Models}

As shown in the example in Table \ref{tab:common_law}, judges often express a legal argument and then quote precedent in support of the argument.
The data collected from CAP was used to train two models to predict the correct target passage given the local context surrounding this passage in the opinion citing it, as well as the global context from the introduction and conclusion of the opinion, which often contain general background relevant to the case.

This task was treated as a multi-class classification problem and two different models were trained.
First, the pretrained LEGAL-BERT model presented by \citet{chalkidis-etal-2020-legal} was fine-tuned to predict passages.
Second, a simple three-layer feed forward neural network (FFNN) using custom continuous bag of word (CBOW) word vectors to represent contexts was trained on the same task.

\subsection{Extracting Local and Global Context from Opinions}

For the FFNN, a local context window of 400 characters to either side of the target passage was extracted to represent the specific legal argument being made.
Additionally, the first and last 2,500 characters in each training opinion were extracted to capture the general opinion context.
To ensure that the word embeddings would successfully capture ``legalese" domain specific meanings of words, a custom 300-dimensional CBOW Word2Vec embedding was trained on all federal judicial opinions (approximately 6 billion tokens).
All tokens that appeared at least 1,000 times were included in the vocabulary.

BERT has a limited input length \citep{devlin2018bert} and so a smaller context window of 300 characters to either side of the target passages as well as 300 characters from the introduction and conclusion were extracted.
LEGAL-BERT is pretrained on legal language and so no further domain specific modifications were made.
The four context windows were concatenated together in the order they appeared in the original opinion to form ``mini-opinions".

Only minimal preprocessing was performed for both models.
All characters other than letters were discarded and all letters were set to lowercase.
The name of and reference to the opinion containing the target passage was removed, if it appeared in the context.
Finally, for the FFNN only, stop words were removed.

\subsection{Balancing Training Data}

\begin{figure}[t!]
\centering
\includegraphics[width=\linewidth]{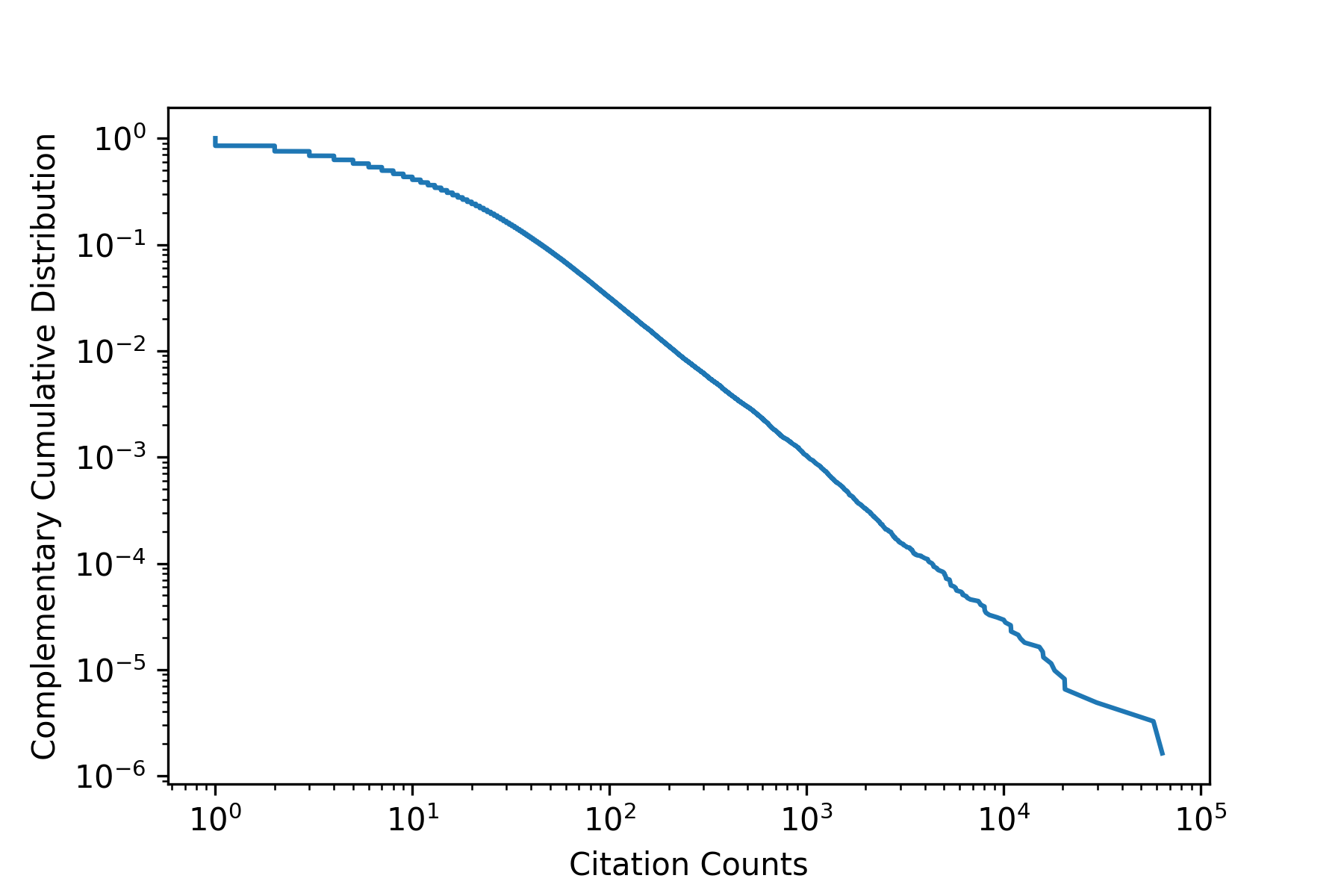}
\caption{Judicial citations obey a power law distribution}
\label{fig:ccdf}
\end{figure}

As shown in Figure \ref{fig:ccdf}, citation frequency obeys a power law distribution, with a small number of citations being cited very frequently.
This distribution results in an imbalanced dataset which must be addressed to avoid training a model that only predicts the most frequent citations.
Moreover, the distribution makes simple under-sampling techniques infeasible: the least cited passage in the training data was cited 26 times, therefore under-sampling would require discarding almost 80\% of the training data.
To overcome these challenges in the FFNN, the minority classes in the training data were synthetically over-sampled, using the synthetic minority over-sampling technique (SMOTE) presented by \citet{chawla2002smote}.
In essence, SMOTE generates new minority class training examples by interpolating between existing minority class data points.

\citet{tayyar-madabushi-etal-2019-cost} suggest that BERT can handle class imbalance with no further data augmentation and so the BERT model was trained on the original imbalanced training data.

\subsection{BERT Model}

Each ``mini-opinion" was tokenized using the LEGAL-BERT tokenizer and padded to a length of 340 tokens (the length of the longest ``mini opinion" was 336 tokens).
The pretrained LEGAL-BERT model was fine-tuned mostly following the suggestions proposed by \citet{devlin2018bert} with the exception of the batch size which was reduced to 8 (instead of the recommended 16 or 32) to enable fine-tuning on a single GeForce RTX™ 2080 GPU with 8GB of memory.
The BERT model was trained to output a 5,000-dimensional vector corresponding to the 5,000 possible passages.
The training data was shuffled and 5\% (28,000 examples) was reserved for testing.

\subsection{FFNN Model}

The input to the FFNN model was the concatenation of four vectors representing the global context at the beginning and end of the opinion, as well as the local contexts before and after each target passage.
Each token in the contexts was individually embedded using the legal CBOW embedding and the embeddings were averaged over each context window.
As before, the target citations were represented as a 5,000-dimensional vector.

The input vectors were fed into a first fully connected hidden layer with a rectified linear (ReLU) activation function and the resulting 512-dimensional vector was fed to a second fully connected hidden layer with a ReLU activation function to generate a 128-dimensional vector which was finally fed to a fully connected output layer with a softmax activation function to generate a 5,000-dimensional output vector.
All three layers used l-2 regularization with a regularization constant of $10^{-3}$.
Batch normalization was applied between the first two hidden layers.
The FFNN model was trained using mini-batches of size 32 for a total of 10 epochs using the ADAM optimizer and categorical cross entropy loss.
No further hyper parameter tuning was performed.
As before, 5\% of the training data was retained for testing.

\section{Results}

\subsection{Macro-averaged evaluation metrics}

The accuracy of multi-class classification models can be misleading, especially when classes are imbalanced as in this case.
\citet{damaschk-etal-2019-multiclass} present a set of evaluation metrics for multi-class text classifications, which have been adapted to evaluate the models' performances.

A correct label $y_{x}$ is associated with each sample $x$ in the test set $X$.
The model predicts a label $\hat y_{x}$ which can be used to calculate the class-wise precision ($P$), recall ($R$) and F-Score ($F$):

\begin{equation}
P(y) = \frac{|\{x \in X | y_{x} = y \wedge  y_{x} = \hat y_{x}\}|}{|\{x \in X | \hat y_{x} = y \}|}
\end{equation}

\begin{equation}
R(y) = \frac{|\{x \in X | y_{x} = y \wedge  y_{x} = \hat y_{x}\}|}{|\{x \in X | y_{x} = y \}|}
\end{equation}

\begin{equation}
F(y) = \frac{2 \cdot P(y) \cdot R(y)}{P(y) + R(y)}
\end{equation}

\begin{center}
\begin{table}
\begin{tabularx}{\linewidth}{|cXccc|}
\hline
Model && $P$ & $R$ & $F$\\ 
\hline
\multirow{2}{4em}{BERT}&Macro& $0.58$ & $0.60$ & $0.56$  \\  
&Weighted & $0.72$ & $0.73$ & $0.71$  \\  
\hline
\multirow{2}{4em}{FFNN}&Macro& $0.37$ & $0.38$ & $0.34$  \\  
&Weighted & $0.39$ & $0.35$ & $0.33$  \\  
\hline
\end{tabularx}
\caption{The fine-tuned BERT model outperforms the FFNN across all metrics but is biased towards the over-represented passages while the FFNN has a comparable performance on both frequent and infrequent passages.}
\label{tab:macro}
\end{table}
\end{center}

For each metric above, two evaluation metrics over the entire unseen test set can be calculated.
First, using the frequencies of the labels as weights, a weighted mean on the metrics can be calculated to obtain a metric that reflects class imbalance.
Second, the three metrics can be macro-averaged, which represents an unweighted mean of each metric, to obtain a measure that is not biased by a strong performance in the most frequent classes.

Evaluating the models on the unseen test set (Table \ref{tab:macro}) shows that the fine-tuned BERT model outperforms the FFNN but that the BERT model is biased towards predicting more frequently cited passages.

\subsection{Top-k Accuracy}

In practice, lawyers would likely use LPP models to make numerous recommendations and use their legal training to decide which of these recommendations should be used in their legal argument.
Therefore, it is sensible to evaluate how often the target passage is among the Top-k predictions produced by the model.

\begin{center}
\begin{table}[h]
\begin{tabularx}{\linewidth}{|X|cccc|}
\hline
Model & Top-10 & Top-20 & Top-50 & Top-100\\ 
\hline
BERT & $0.96$ & $0.98$ & $0.99$ & $0.99$  \\  
\hline
FFNN & $0.81$ & $0.90$ & $0.95$ & $0.98$  \\  
\hline
\end{tabularx}
\caption{Although both models reliably predict the correct passage, the BERT model outperforms the FFNN's Top-k accuracy for all values of k between 10 and 100.}
\label{tab:topk}
\vspace{-8mm}
\end{table}
\end{center}

Comparing the Top-k accuracy for both models on the unseen test set (Table \ref{tab:topk}) shows that both are able to place the correct citation among the top predictions, although the BERT model clearly outperforms the FFNN.

\subsection{Evaluation Summary}
The Top-k accuracy an important metric since it is especially useful for lawyers in practice.
The reasonably high Top-k accuracy indicates that both models have successfully learned the semantic connection between legal arguments and relevant citation passages.

The BERT model, unlike the FFNN, is biased towards predicting the more frequently cited passages, however, it outperforms the FFNN across all other metrics.
It must be underscored that the BERT model is much more resource intensive to train.
Both models were trained on the same GPU and while the FFNN (1,327,624 parameters) was trained in 13 minutes, the BERT model (113,327,240 parameters) required 20 hours of training time.

Given the skill and training associated with legal practice, these results are highly encouraging and suggest that NLP techniques have the ability to predict precedent relevant to legal arguments, at least when these arguments rest on commonly cited precedent.
The results suggest a perhaps unsurprisingly close semantic connection between the context of a legal argument (both local and global) and the relevant citations.
The models demonstrate that identifying the correct legal precedent on the basis of semantic context alone is possible, and no extraneous legal knowledge is required in many instances.

\section{LPP in the Wild}

The unseen testing data, like the training data, came from judicial opinions, in practice however, LPP models are likely to be used by lawyers to draft briefs more efficiently and effectively.
In contrast to judicial opinions, legal briefs cannot be readily obtained en masse in a machine readable format.
Therefore, two complex high quality legal briefs were chosen at random to provide an insight into model performance in a realistic setting.
Due to its superior performance, only the BERT model was evaluated on the briefs.

\begin{table*}[h!t]
\centering
\begin{tabularx}{0.955\textwidth}{|c|l|}
  \hline
  \multicolumn{2} { | p{0.927\textwidth}|  }{\textbf{Model Input: }\textit{The district court did not err in denying petitioner’s motions for a change of venue. Petitioner’s central contention is that the degree and nature of pretrial publicity, and the impact of Enron’s collapse on Houston, gave rise to an irrebuttable presumption of prejudice among the entire venire. But this Court’s cases — and the Constitution — are satisfied if the jurors who actually sat are impartial. Because the jury that decided his case was impartial, petitioner has failed to establish a constitutional violation. The Constitution requires trial before an impartial jury. The district court’s denial of petitioner’s motion for a change of venue did not violate his constitutional rights.}} \\
  \hline
  \hline
  Proposed Passage & Relevance \\
  \hline
  \multicolumn{1}{ | p{0.7\linewidth}|}{This Court has long held that the remedy for allegations of juror partiality is a hearing in which the defendant has the opportunity to prove actual bias. \textit{Smith v. Phillips}, 455 U.S. 209, 215 (1982)} & Relevant; Cited \\
  \hline
  \multicolumn{1}{ | p{0.7\linewidth}|}{In essence, the right to jury trial guarantees to the criminally accused a fair trial by a panel of impartial, “indifferent” jurors. \textit{Irvin v. Dowd}, 366 U.S. 717, 722 (1961)} 
  & \multicolumn{1}{ | p{0.2\linewidth}|}{Relevant; Another passage cited} \\
  \hline
  \multicolumn{1}{ | p{0.7\linewidth}|}{Due process means a jury capable and willing to decide the case solely on the evidence before it. \textit{Smith v. Phillips}, 455 U.S. 209 (1982)} & Relevant; Cited \\
  \hline
  \multicolumn{1}{ | p{0.7\linewidth}|}{The voir dire in this case indicates no such hostility to petitioner by the jurors who served in his trial as to suggest a partiality that could not be laid aside. \textit{Murphy v. Florida}, 421 U.S. 794, 800 (1975).} 
  & \multicolumn{1}{ | p{0.2\linewidth}|}{Relevant; Another passage cited} \\
  \hline
  \multicolumn{1}{ | p{0.7\linewidth}|}{It is sufficient if the juror can lay aside his impression or opinion and render a verdict based on the evidence presented in court. \textit{Irvin v. Dowd}, 366 U.S. 717, 723 (1961)} 
  & Relevant; Cited \\
  \hline
  \multicolumn{1}{ | p{0.7\linewidth}|}{The test under \textit{Kotteakos} is whether the error “had substantial and injurious effect or influence in determining the jury’s verdict.” \textit{Brecht v. Abrahamson}, 507 U.S. 619, 637 (1993)} & Relevant; Not Cited \\
  \hline
  \multicolumn{1}{ | p{0.7\linewidth}|}{In a criminal case, any private communication, contact, or tampering, directly or indirectly, with a juror during a trial about the matter pending before the jury is, for obvious reasons, deemed presumptively prejudicial, if not made in pursuance of known rules of the court and the instructions and directions of the court made during the trial, with full knowledge of the parties. \textit{Remmer v. United States}, 347 U.S. 227, 229 (1954)} & Relevant; Not Cited \\
  \hline
\end{tabularx}
\caption{Seven of the ten predicted citations are relevant to the argument in the brief.}
\label{tab:Kagan}
\end{table*}

The first brief \citep{kaganbrief} was submitted to the U.S. Supreme Court.
It was co-authored by Elena Kagan, a former U.S. Solicitor General and current U.S. Supreme Court Justice. 
The brief is 86 pages long and, among other things, argues that the Petitioner's right to a fair trial was not violated due to the pretrial publicity his case had received. 
Five sentences that seemed to provide a good overview of the argument were manually extracted from the brief and used as an input to the BERT model with no further preprocessing. 
The top-10 predictions from the model were subsequently analyzed to determine whether (a) they had appeared in the brief, and (b) they were relevant to the argument made in the brief. 
Of the ten predicted passages, seven were deemed relevant to the brief. 
Table \ref{tab:Kagan} shows these seven relevant passages, of which three appeared in the brief, two belonged to an opinion from which another passage was cited in the brief, and two seemed relevant but were not cited.

\begin{table*}[h!t]
\centering
\begin{tabularx}{0.955\textwidth}{|c|l|}
  \hline
  \multicolumn{2} { | p{0.927\textwidth}|  }{\textbf{Model Input: }\textit{More than 100 years ago, the Supreme Court ruled that the protection of our Constitution reaches all people within the territory of the United States regardless of their citizenship. The Fourteenth Amendment’s pledges of due process and equal protection apply. Congress’s “Plenary Power” over immigration is the source of the entry fiction. The entry-fiction doctrine derives from the recognition that the political branches of government are more appropriately suited to function as gatekeeper of the nation’s borders. Even if plenary, Congress’s immigration power has never been unlimited. The Court should affirm the district court’s denial of summary judgment because the Fourth and Fifth Amendments did protect Martinez-Agüero from gross physical abuse and inhumane treatment once she entered the territory of the United States.}} \\
  \hline
  \hline
  Proposed Passage & Relevance \\
  \hline
  \multicolumn{1}{ | p{0.7\linewidth}|}{It is well established that the Fifth Amendment entitles aliens to due process of law in deportation proceedings. \textit{Reno v. Flores}, 507 U.S. 292, 306 (1993).”} & Relevant; Not Cited \\
  \hline
  \multicolumn{1}{ | p{0.7\linewidth}|}{Courts have long recognized the power to expel or exclude aliens as a fundamental sovereign attribute exercised by the Government’s political departments largely immune from judicial control. \textit{Shaughnessy v. United States ex rel. Mezei}, 345 U.S. 206, 210 (1953).} & Relevant; Cited \\
  \hline
  \multicolumn{1}{ | p{0.7\linewidth}|}{The Court held that “when the Executive exercises this [delegated] power negatively on the basis of a facially legitimate and bona fide reason, the courts will neither look behind the exercise of that discretion, nor test it by balancing its justification against the First Amendment interests of those who seek personal communication with the applicant. \textit{Fiallo v. Bell}, 430 U.S. 787, 794 (1977)} 
  & \multicolumn{1}{ | p{0.2\linewidth}|}{Relevant; Another passage cited} \\
  \hline
  \multicolumn{1}{ | p{0.7\linewidth}|}{This Court has repeatedly emphasized that “over no conceivable subject is the legislative power of Congress more complete than it is over” the admission of aliens. \textit{Fiallo v. Bell}, 430 U.S. 787, 792 (1977)} 
  & \multicolumn{1}{ | p{0.2\linewidth}|}{Relevant; Another passage cited} \\
  \hline
  \multicolumn{1}{ | p{0.7\linewidth}|}{We hold that when the Executive exercises this power negatively on the basis of a facially legitimate and bona fide reason, the courts will neither look behind the exercise of that discretion, nor test it by balancing its justification against the First Amendment interests of those who seek personal communication with the applicant. \textit{Kleindienst v. Mandel}, 408 U.S. 753, 770 (1972)} 
  & \multicolumn{1}{ | p{0.2\linewidth}|}{Relevant; Another passage cited} \\
  \hline
  \multicolumn{1}{ | p{0.7\linewidth}|}{Our cases “have long recognized the power to expel or exclude aliens as a fundamental sovereign attribute exercised by the Government's political departments largely immune from judicial control. \textit{Fiallo v. Bell}, 430 U.S. 787, 792 (1977)} 
  & \multicolumn{1}{ | p{0.2\linewidth}|}{Relevant; Another passage cited} \\
  \hline
  \multicolumn{1}{ | p{0.7\linewidth}|}{For a host of constitutional and statutory provisions rest on the premise that a legitimate distinction between citizens and aliens may justify attributes and benefits for one class not accorded to the other; and the class of aliens is itself a heterogeneous multitude of persons with a wide-ranging variety of ties to this country. \textit{Mathews v. Diaz}, 426 U.S. 67, 79 (1976).} 
  & \multicolumn{1}{ | p{0.2\linewidth}|}{Relevant; Another passage cited}
  \\
  \hline
\end{tabularx}
\caption{Seven of the ten predicted citations are relevant to the brief. Although several predicted passages stem from the same opinion, these passages themselves contain relevant references to other opinions, several of which appear in the legal brief. }
\label{tab:SG}
\end{table*}

The second brief \citep{SGbrief} was featured in \citet{garner2014winning} as one of the ``best ever" legal briefs.
The 62-page brief argued that an alien stopped at a U.S. border has a constitutional right to be free from false imprisonment and from the use of excessive force by U.S. law enforcement personnel.
As before, the BERT model was given an input consisting of a few sentences from the brief and generated ten predictions.
Table \ref{tab:SG} shows the seven predictions that were deemed relevant to the brief.
One was cited in the brief, five belonged to an opinion from which another passage was cited, and one seemed relevant but was not cited.
The model provided three passages from the same opinion and each of these passages cited a passage in another opinion which was relevant to the brief.

Overall, the performance on these legal briefs provides an indication that the BERT LPP model would perform well in practice.
Several highly skilled attorneys spent hundreds of hours working on these briefs and the fact that a model can predict relevant passages of precedent based on a few sentences of context is highly encouraging.
This section cannot show definitively whether the BERT model's bias towards more frequently cited passages poses a practical problem.
However, the two examined briefs are very distinct, and the model made relevant predictions in both cases suggesting that the bias may not be a problem in practice.

\section{Ethical Considerations}

The work outlined in this paper has the potential to augment attorneys' ability to identify relevant precedent in a cost- and time-effective manner, thereby increasing access to justice.

That being said, given the high stakes inevitably involved in litigation, it is especially important to consider the ramifications of using automated means to identify precedent.
The following discussion will highlight two such ramifications, but this brief discussion is no substitute for a full exploration of the social and ethical considerations related to automating legal research, an analysis beyond the scope of this paper.

The models presented here are trained exclusively on prior citations generated by judges in the past.
Therefore, any biases present in these citations would inevitably be learned by the model.
It is generally accepted that a number of factors (including ideology, workload and public perception) have the potential to influence judicial reasoning \citep{landes1976legal}.
While there is a natural tendency to cite previously cited precedent, lawyers have the ability to cite new portions of opinions and this ability is a fundamental requirement for the continued evolution of the law.
If automated legal research techniques become ubiquitous, this has the potential to not only perpetuate historical biases but also to cement the legal status quo and make change less probable.

It is likely that a highly skilled attorney, using the resources provided by legal research platforms and working without time constraints, will more successfully identify precedent than an automated approach alone.
In practice, it is likely that attorneys would use LPP to help them narrow their search and economize their time.
Nonetheless, parties to legal disputes who must rely primarily on automated research tools will be at a disadvantage compared to those who can also afford to do so manually.
It is conceivable that, when automated tools perform poorly or make mistakes a human lawyer would be unlikely to make, this could seriously jeopardize a party's standing in court.
While automated legal research methods have the potential to alleviate widespread access to justice issues, a thorough understanding of their likely failure modes is required to ensure that lawyers know when relying on these tools alone may not be appropriate.

As demonstrated here, the law is a promising area for the deployment of NLP tools and these tools have the potential to improve access to justice.
However, any such endeavor must be pursued thoughtfully and carefully, lest well intentioned tools do more harm than good.
Moreover, successful implementation of NLP in the law requires an exchange of ideas between NLP researchers and legal professionals to ensure that the tools that are developed address real issues and that the law is aware of the potential and limitations of these tools.

\section{Future Work}

In many ways, the law is a discipline built on language yet it has done well to evade the scrutiny of computational linguists.
This section will outline a few avenues for future research based on the models presented here, although there are undoubtedly countless others.

A weakness of the proposed model is that it is confined to predicting frequently cited passages.
While this reflects how most lawyers tend to operate, it comes with the drawbacks discussed in the Ethical Considerations section.
A simple improvement would be to include more passages among the model's possible outputs.
A more sophisticated approach would be to train a model to generate a representation of a relevant citation rather than a categorical prediction.
Such a model could be used to identify relevant passages across all opinions, regardless of whether they have been previously cited, by identifying the passage that is closest in embedding space to the predicted output.
Repeatedly calculating minimum distances in embedding space over hundreds of millions of possible passages is unlikely to be computationally tractable and so this approach would likely hinge on an efficient initial narrowing of candidate passages.

The current model's input is a simple summary of the legal argument being made.
However, a similar approach could be used to test the quality of legal arguments by adding a set of citations to the input to predict any missing relevant citations.
Such a tool could be useful to lawyers who may worry that they missed an important piece of precedent, or to judges who wish to ascertain whether the legal briefs before them fail to address important legal concepts.

\section{Conclusion}

Judicial precedent is the bedrock of legal reasoning in common law jurisdictions such as the United States.
This paper introduces LPP, the task of predicting relevant passages from precedential court decisions given the context of a legal argument.
Through the LPP task, this paper demonstrates how NLP can be used to address an unmet need of the legal community and increase access to justice.
LPP can augment attorneys' ability to identify precedent in a cost- and time-efficient manner to reduce the cost of litigation and accordingly increase access to justice.
This paper showcases a BERT model trained on 530,000 examples of U.S. federal judges advancing legal arguments by citing precedent.
In 96\% of unseen test examples the correct passage is among the top-10 predicted passages.
The same model can predict relevant precedent given a short summary of a complex and unseen legal brief, predicting the precedent that was actually cited by the brief's co-author, former U.S. Solicitor General and current U.S. Supreme Court Justice Elena Kagan.

\bibliographystyle{AutoLaw_natbib}
\bibliography{AutoLaw}


\end{document}